\documentclass{article}
\usepackage{graphicx}
\usepackage{booktabs}
\usepackage{amsmath,amsfonts}
\usepackage{algorithmic}
\usepackage{algorithm}
\usepackage{array}
\usepackage[caption=false,font=normalsize,labelfont=sf,textfont=sf]{subfig}
\usepackage{textcomp}
\usepackage{stfloats}
\usepackage{url}
\usepackage{verbatim}
\usepackage{graphicx}
\usepackage{cite}
\usepackage{comment}
\usepackage{multirow, array}
\usepackage{makecell}
\usepackage{pifont}
\usepackage{svg}
\usepackage{float}
\usepackage{authblk}

 \date{}

\begin{document}

\title{Beyond Generative Artificial Intelligence: Roadmap for Natural Language Generation}

\author[]{María Miró Maestre}
\author[]{Iván Martínez-Murillo}
\author[]{Tania J. Martin}
\author[]{Borja Navarro-Colorado}
\author[]{Antonio Ferrández}
\author[]{Armando Suárez Cueto}
\author[]{Elena Lloret}
\affil[]{Department of Software and Computing Systems (DLSI), University of Alicante, Spain}

\maketitle

\begin{abstract}
Generative Artificial Intelligence has grown exponentially as a result of Large Language Models (LLMs). This has been possible because of the impressive performance of deep learning methods created within the field of Natural Language Processing (NLP) and its subfield Natural Language Generation (NLG), which is the focus of this paper. Within the growing LLM family are the popular GPT-4, Bard and more specifically, tools such as ChatGPT have become a benchmark for other LLMs when solving most of the tasks involved in NLG research. This scenario poses new questions about the next steps for NLG and how the field can adapt and evolve to deal with new challenges in the era of LLMs. 
To address this, the present paper conducts a review of a representative sample of surveys recently published in NLG. By doing so, we aim to provide the scientific community with a research roadmap to identify which NLG aspects are still not suitably addressed by LLMs, as well as suggest future lines of research that should be addressed going forward. 
\end{abstract}

\section{Introduction}
Natural Language Processing (NLP) is a key component of Artificial Intelligence (AI) in the sense that it enables humans and machines to interact more naturally. Despite NLP's recent popularity, 
research in this area spans more than 60 years. The complexity involved in the understanding ---Natural Language Understanding (NLU)---  and the production of languages ---Natural Language Generation (NLG)--- is evident in the relatively limited performance of more semantic and pragmatic tasks, such as Word Sense Disambiguation, Coreference Resolution, or Intention Detection. 

Since the emergence of Large Language Models (LLMs) and the so-called Generative AI, there has been an exponential growth of different LLMs families (GPT, Bert, BLOOM, LLaMa, etc.) \cite{zhao2023survey}. Moreover, several AI tools have been developed, with ChatGPT being the most popular. Indeed, ChatGPT has revolutionized the way information is automatically generated and provided to the user.

Until very recently, 
AI systems were focused on specific tasks, such as Question Answering, Description Generation, or Text Summarization. However, LLMs are trained over tons of information, making it possible for a single NLG system to address many applications, i.e., following a one-fits-all approach. This is the case, for instance, of ChatGPT, which was originally conceived as a chatbot, although it now provides solutions in natural language to a wide range of prompts (open questions, poetry generation, summaries, etc.). 
The popularity of these NLG tools, partly because of their versatility in the variety of tasks they solve, has placed AI research on the radar, in particular NLP. 

Indeed, great advances have been made in NLP tasks thanks to neural models and the aforementioned LLMs (as 
machine translation, text classification, and text generation). The progress has been so great that some of these tasks can now be considered solved. The question arises as to how this will impact NLP and NLG going forward and how will their role shift in the face of recent advances in LLMs. 

Languages are, however, more complex and ultimately LLMs are only specific models based mainly on contextual relationships between words. Indeed, new tasks or new 
NLU and NLG research lines are emerging, and others remain unsolved. Papers as \cite{ChurchLiberman2021} indicate some of the unsolved topics, such as syntactic parsing with Universal Dependencies, semantic compositionality or causality relationships.

The overall goal of this survey is to provide an analysis of several NLG survey papers published recently, exploring the emerging and unsolved research topics in NLG. Our work is presented as a NLG roadmap, detecting the areas requiring improvement and looking beyond the recent successes of Generative AI tools. We consider this to be of value to the research community in terms of revealing the key areas that need to be tackled in NLG going forward. 

The paper is organized as follows. First, we introduce how NLG has evolved throughout the years in the ``NLG Evolution'' section. Then, we move on to describe the approach chosen to compile the different surveys that structure this review in the section ``Methodology''. The NLG survey review together with Table \ref{table_surveys} showing the aspects covered in each survey is explained in the section ``Analysis of the surveys''. We define our proposal for a NLG research roadmap in the ``Identified research gaps section'', where we state which research issues should be addressed in the NLG discipline to improve the performance of current LLMs. The next section, ``New considerations triggered by generative AI'', goes one step further on the roadmap, identifying which lines of research should receive attention as a consequence of the latest models in generative AI. Finally, in the ``Conclusions'' section, we briefly restate the identified roadmap with a view to highlighting the many different tasks that need to be studied within the field.

\section{NLG Evolution}
The NLG field has changed drastically from when it was first studied in the end of the 1970s \cite{mcdonald2010natural}. Originally, NLG architectures were a sequential pipeline of the following three well-differentiated stages: (1) Macroplanning, which determines what content to include in the final output; (2) Microplanning, which establishes how to include the selected content and (3) Realization, which generates the final output with complete meaning. All the architectures following this pipeline structure are known as modular architectures, with the standard architecture being the one proposed in \cite{reiter1994consensus}. Afterwards, these stages became more flexible, giving rise to a new approach known as planning perspectives. Although this group of architectures still contemplated a task division, that division was less strict than in modular architectures, enabling two or more different tasks to be combined and performed as one step. 

Finally, task division started to disappear, and was replaced by what is now defined as global approaches. These architectures rely on statistical learning and perform the generation in just one stage. The major milestone within this group was the Transformer architecture, which achieved great results on NLG tasks \cite{topal2021transformers}. Since then, several architectures based on Transformers have been proposed, with LLMs delivering better results and producing texts almost indistinguishable from texts written by humans. 

Nowadays, the research scope in NLG is focused on developing larger LLMs, which are neural networks with billions of parameters. 
Although these models achieve an impressive performance in generation tasks, they still lack precision and have some problems in generating texts faithfully in the same way that humans do, as we will argue in the next sections.

\section{What Recent Survey Studies say?}
Recapping surveys on the current state of NLG holds significant importance in understanding and assessing developments in this broad and evolving field. In this section, we will discuss the employed methodology for our survey compilation and examine the key findings and trends that emerge from the examined surveys. By analyzing these surveys, we aim to get a more comprehensive insight into the current and future directions of natural language generation research.

\subsection{Methodology}
The methodology used to gather the NLG surveys that shape our analysis was to first review the exhaustive selection of NLG surveys included in \cite{marta2021discourse}. From this starting point and by broadening our search to more updated publications, we decided which NLG aspects such surveys should include to cover the different research approaches shown in the field. Consequently, a first determinant for filtering our search was finding surveys covering studies that ranged from chronological perspectives of the evolution of NLG systems to theoretical reviews of the state of the art regarding both traditional and neural models. Moreover, we made sure that some of those surveys dedicated part of their analysis to the evolution of the main techniques used for core tasks in NLG, as well as evaluation methods and current issues that need attention. 

Once we gathered the surveys that met the selection criteria, we limited the time interval to works published from 2016 to 2023 to ensure the relevance of the tasks that the research community is currently exploring and the gaps identified in the field. With this in mind, we created a corpus of 16 NLG surveys which generally cover the following research aspects:

\begin{itemize}

   \item{\it Main objectives and contributions}---We first determined the aim and the core strengths of each survey to compare traditional approaches to those focused on the explosion of the latest neural models, as well as to check what new knowledge they provide to the research community.

    \item{\it Inclusion of corpora}---One of the crucial elements of an NLG system is a corpus to apply the chosen architecture. Consequently, we 
    also verified which tasks have more datasets available and for what language.

    \item{\it Methods}---Another central factor to consider 
    was the different techniques used for creating NLG architectures, to check which scientific approaches are adopted nowadays and for which languages, to test and improve the validity of the systems.

    \item{\it Tools}---Similarly, checking which surveys included tools or demos of NLG software that proved the effectiveness of the described architectures was useful for deducing what final applications and languages are trending areas in the discipline.

    \item{\it Conclusions}---Finally, our review methodology helped us detect research gaps that should be addressed in future; indeed, some of these are flagged in the conclusions of the surveys.

\end{itemize}

\subsection{Survey Exploration}

Table \ref{table_surveys} gathers data on the year the survey was published, and whether the survey includes the following: corpora, methods, and tools. The table also interprets the data, presenting some descriptive statistics to indicate research gaps and thereby, opportunities.

\begin{table}[!]
\caption{Analysis of NLG surveys}
\label{table_surveys}
\begin{tabular}{p{7cm}p{1cm}p{1.5cm}p{1.5cm}p{1.5cm}}
\toprule
\textbf{Survey} & \textbf{Year} & \textbf{Corpora} & \textbf{Methods} & \textbf{Tools} \\
A survey of natural language generation \cite{dong2023surveygen} & 
2023 & \ding{52} & \ding{52} & \ding{52}  \\[3pt]
Survey of hallucination in natural language generation \cite{ji2023survey} & 
2023 & \ding{52} & \ding{52} & \ding{54}  \\[3pt]
A survey of knowledge-enhanced text generation \cite{yu2022textgeneration} & 
2022  & \ding{52} & \ding{52} & \ding{52} \\[3pt]
Neural natural language generation: A survey on multilinguality, multimodality, controllability and learning \cite{erdem2022neural} & 
2022 & 
\ding{52} & \ding{52} & \ding{54}  \\[3pt]
Recent advances in neural text generation: A task-agnostic survey \cite{tang2022survey} & 
2022 & 
  \ding{52} & \ding{52} & \ding{54}  \\[3pt]
The survey: Text generation models in deep learning \cite{iqbal2022survey} & 
2022 & \ding{54} & \ding{52} & \ding{54}  \\[3pt]
Exploring transformers in natural language generation: GPT, BERT, and XLNet \cite{topal2021transformers} & 2021 & \ding{54} & \ding{52} & \ding{54}  \\[3pt]
Positioning yourself in the maze of neural text generation: A task-agnostic survey \cite{chandu2021survey} & 2021 & \ding{54} & \ding{52} & \ding{54} \\[3pt]
Automatic story generation: Challenges and attempts \cite{alabdulkarim2021} & 2021 &  \ding{52} & \ding{52} & \ding{52} \\[3pt]
Natural language generation: The commercial state of the art in 2020 \cite{dale2020generation} & 2020 & \ding{54} &  \ding{54}  & \ding{52} \\[3pt]
A survey of natural language generation techniques with a focus on dialogue systems - past, present and future directions \cite{santhanam2019dialogue} & 
2019 & \ding{52} & \ding{52}  & \ding{52}  \\[3pt]
Survey of the state of the art in natural language generation: Core tasks, applications and evaluation
\cite{gatt2018surveynlg} & 2018 & \ding{52} & \ding{52} & \ding{52} \\[3pt]
Neural text generation: Past, present and beyond \cite{lu2018neural} & 2018 & \ding{52} & \ding{52} & \ding{52}  \\[3pt]
A survey on intelligent poetry generation: Languages, features, techniques, reutilization and evaluation \cite{gonçalo2017survey} & 2017 & 
\ding{54} & \ding{52} & \ding{52} \\[3pt]
Recent advances in natural language generation: A survey and classification of the empirical literature \cite{perera2017survey} & 
2017  & \ding{52} & \ding{52} & \ding{52} \\[3pt]
A survey on story generation techniques for authoring computational narratives \cite{kybartas2016survey} & 
2016 & 
\ding{54} & \ding{54} & \ding{52}  \\[3pt]
\end{tabular}
\end{table}

At a more macro level, the  16 papers analyzed can be grouped into three main categories. The first group focuses on providing an overview of the NLG field with an indication of future research directions. Santhanam and Shaikh \cite{santhanam2019dialogue} provide a comprehensive overview of NLG approaches and suggest avenues for future research in open domain dialogue systems. Gatt and Krahmer \cite{gatt2018surveynlg} explore developments in NLG since 2000, with a focus on data-driven techniques, vision-to-text generation, and the generation of artistic texts. Dale \cite{dale2020generation} specifically examines commercial applications of NLG software, while also presenting an up-to-date overview and discussing challenges and limitations of using NLG in contexts such as non-English languages and highly technical domains. Yu \textit{et al.} \cite{yu2022textgeneration} present a comprehensive review of the work done in the field of knowledge-enhanced text generation. 

The second group of papers provides a holistic overview of advancements in Neural Natural Language Generation (NNLG), a recent and growing research field. Erdem \textit{et al.} \cite{erdem2022neural} investigate recent developments and applications of NNLG from a multidimensional perspective, such as multimodality, multilinguality, controllability and learning strategies. 
Tang \textit{et al.} \cite{tang2022survey} conduct a comprehensive survey of recent advancements in NNLG, categorising them into data construction, neural frameworks, training strategies, and evaluation metrics. Lu \textit{et al.} \cite{lu2018neural} systematically survey NNLG, comparing properties of the models and their techniques through benchmarking experiments. Topal \textit{et al.} \cite{topal2021transformers} focus on deep generative modelling for text generation, considering papers from 2015 onwards and evaluating approaches in different application domains. Chandu and Black \cite{chandu2021survey} offer a task-agnostic survey of modelling approaches in neural text generation, assisting researchers in positioning their work and identifying new challenges. 
Iqbal and Qureshi \cite{iqbal2022survey} review various deep learning models used for text generation explaining the progress made in this area. 

The third group concentrates on specific areas or tasks within NLG. Perera and Nand \cite{perera2017survey} offer a detailed overview and classification of state-of-the-art approaches in NLG, particularly related to document planning, micro-planning, and surface realisation modules. Kybartas and Bidarra \cite{kybartas2016survey} examine the automated versus manual authoring of plot and space components in story generation. Gonçalo \cite{gonçalo2017survey} surveys intelligent poetry generators, focusing on languages, form and content features, techniques, reutilization of material, and evaluation. Alabdulkarim \textit{et al.} \cite{alabdulkarim2021} analyze machine learning approaches in story generation, addressing controllability, commonsense knowledge incorporation, reasonable character actions, and creative language generation. Ji \textit{et al.} \cite{ji2023survey} provide a broad overview of the research progress and challenges in the hallucination problem in NLG, covering metrics, mitigation methods, and task-specific advancements in the most common NLG tasks. 
Dhong \textit{et al.} \cite{dong2023surveygen} review NLG research, emphasizing data-to-text and text-to-text generation deep learning methods, as well as new applications, architectures, datasets, and evaluation challenges.

Overall, the 16 papers cover a wide range of topics in NLG, offering insights into commercial applications, deep learning, knowledge integration, evaluation metrics, and specific tasks across various domains. They contribute to understanding the current state of the field and identifying future research directions.

\section{Which Research Gaps Need Attention in NLG?}

\begin{figure*}[h]
\centering
\includegraphics[width=0.9\textwidth]{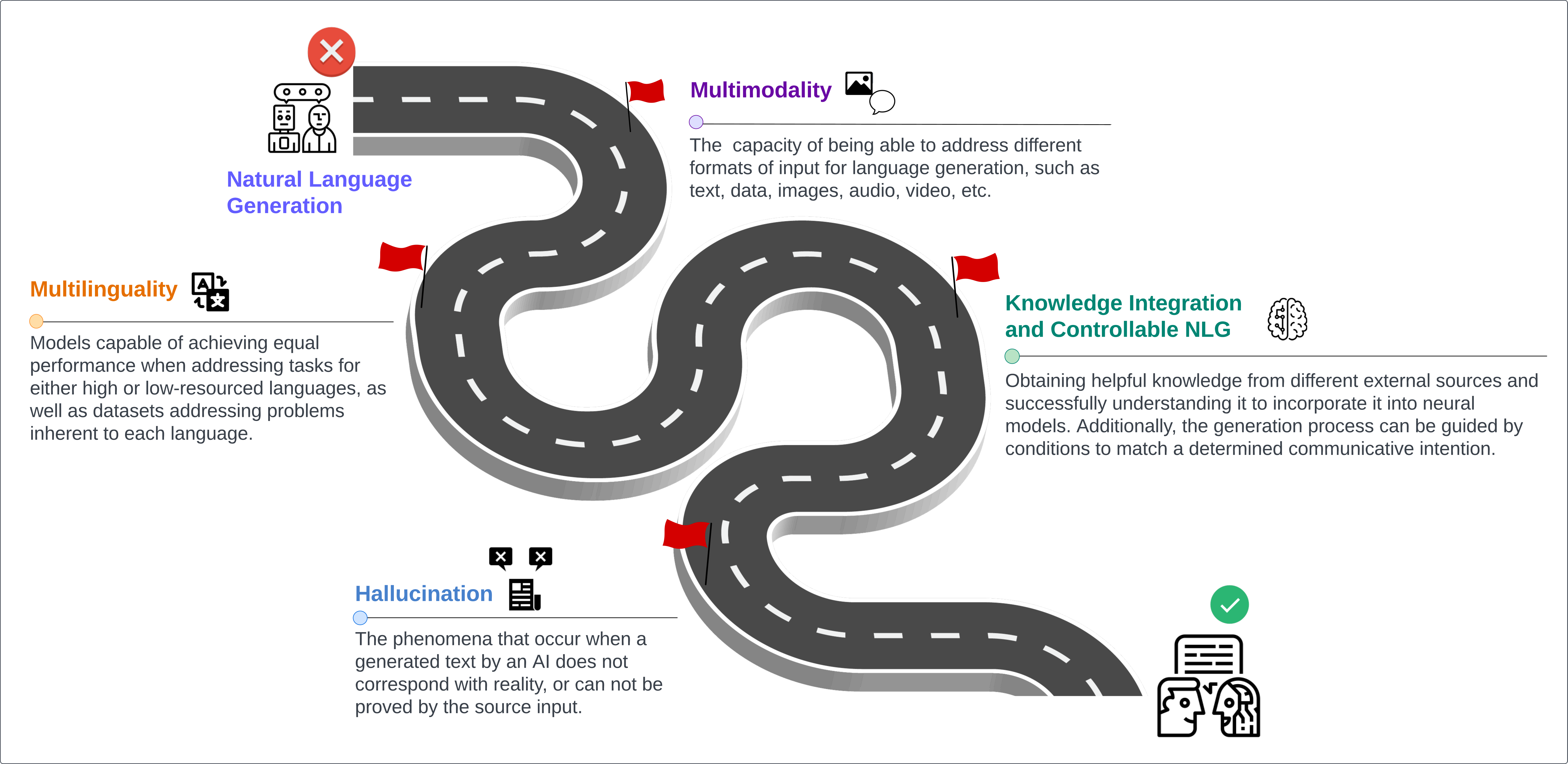}  
\caption{Roadmap of research gaps to address in NLG research.}
\label{fig_roadmap}
\end{figure*}

This survey 
serves as a starting point to identify possible research gaps in NLG tasks, given the broad range of approaches from which these research issues are addressed. The results presented in Table \ref{table_surveys} suggest that the excellent performance of LLMs in several NLG tasks has revolutionized this discipline in an incredibly short time frame of four years (from 2019 with the emergence of GPT3 or T5 up to now). With this immensely rapid development, we now face far more complex tasks that require the input of further contextual knowledge and information modalities in order to achieve a performance that is actually comparable to a text written by a human. 


In line with this, Figure \ref{fig_roadmap} serves as a roadmap of research gaps we have identified through the survey review conducted. Additionally, we checked the existence of each particular issue by using two well-known models, i.e., GPT-4 \cite{OpenAI2023gpt4} through Bing's interface and Google's Bard \cite{manyika2023overview}. In this way, we wanted to verify the current existence of these gaps in two of the latest LLMs available in NLP research to confirm our research findings for the particular case of the Spanish language. By addressing those gaps, we aim to ensure that LLMs cover complex aspects of language that would improve their overall performance for more demanding tasks. The identified research gaps we need to face in the NLG field are the following:

\subsection{Multimodality}

Multimodality refers to the capacity of being able to address different formats of input for language generation, such as text, data, images, audio, video, etc. \cite{erdem2022neural} 
This combined representation of different data formats 
represents an innovative approach to make NLG models improve their contextual knowledge, therefore boosting the addition of commonsense to the generated text, which constitutes one of the issues currently addressed in NLG. Indeed, much research work focuses on this multimodal input format. 
However, most of them tend to prioritize the information given in one of the modalities over the other (either data or text), therefore worsening the balance between the knowledge acquired from each input type \cite{erdem2022neural}. 

\begin{figure}[h!]
\centering
\includegraphics[width=0.7\textwidth]{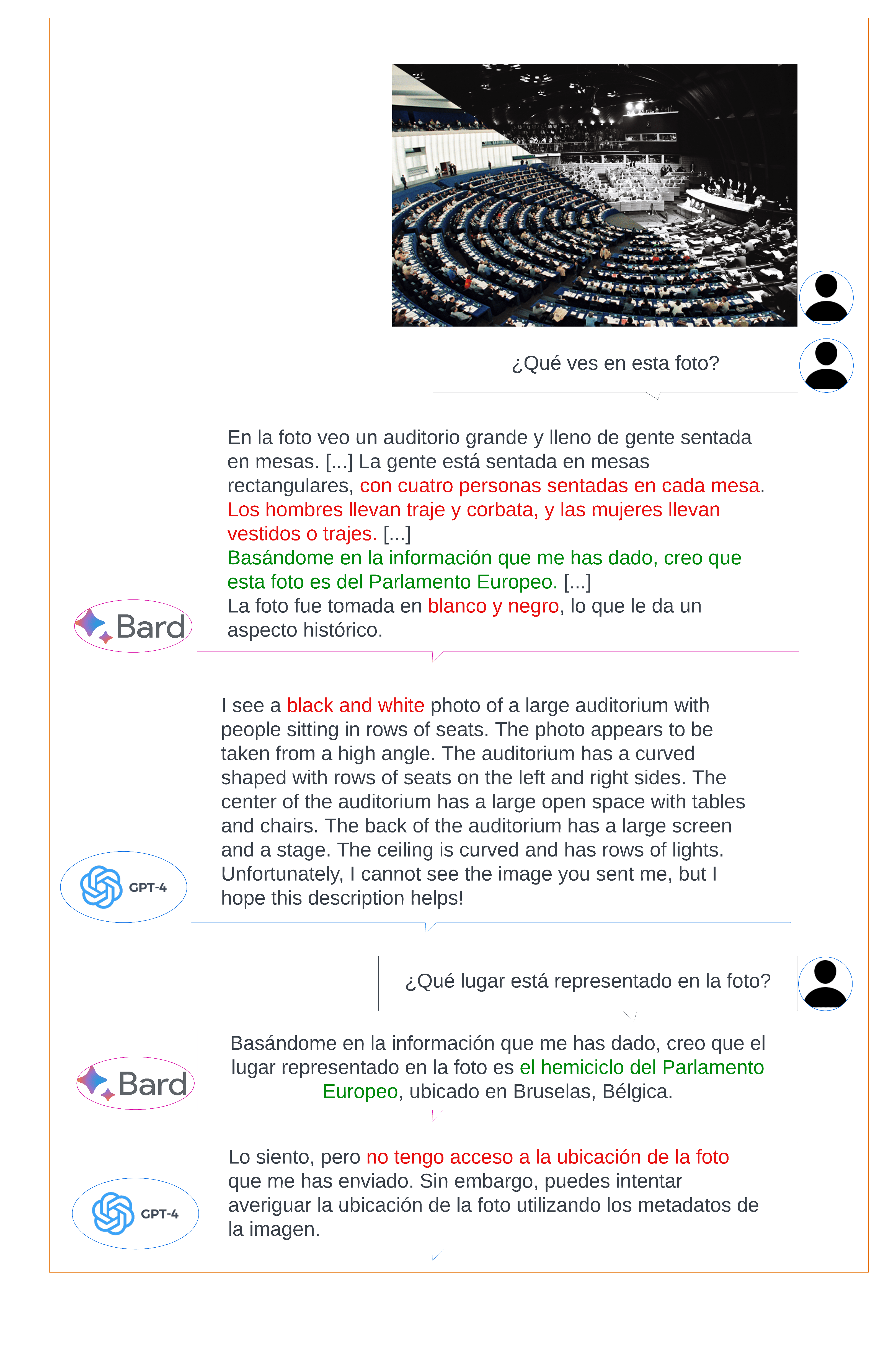}  
\caption{Example of a chat with Bard and GPT-4 with a multimodal input.}
\label{fig_multimodal}
\end{figure}

This need for further improvement when addressing NLG from a multimodal perspective can be seen through simple examples we tested in both Bard and GPT-4 through Bing interface, illustrated in Figure \ref{fig_multimodal}. We showed both models a photograph from an international session at the European Parliament. We did this test in the Spanish version of both chatbots, but GPT-4 first answered in English, whereas Bard directly answered in Spanish. Both GPT-4 and Bard describe the photograph as if it was in black and white, although only half of the picture is actually in those colours. Moreover, when asking each chatbot which place is represented in the photograph, GPT-4 is incapable of assuming that it is one of the plenary rooms of the European Parliament, whereas Bard correctly answers the location without asking for additional information. However, this model also shows examples of hallucination and gender bias issues. On the one hand, it says that there are only 4 people per table, although it is clear there are way more than 4 people. On the other hand, it also states that men wear suits and ties, whereas women are dressed in suits and dresses. Nevertheless, as most of the people in the photograph are sat on their respective seats, we cannot confirm the particular clothes each one is wearing.

Given this inequality when processing the information contained in several modalities, 
we concluded that NLG systems need more multimodal training datasets to improve their performance. In this way, such systems would not miss the extra-linguistic information that may be detected by the combination of several information modalities. Moreover, an additional gap is to evaluate the knowledge balance between such formats to successfully solve some of the many emerging NLG tasks that make use of multimodal datasets, such as speech recognition, visual recognition, machine translation, etc. 

\subsection{Multilinguality}

Multilinguality is another key issue not only for NLG systems but for NLP in general. The Internet has exacerbated the predominance of some languages over others that are at risk of becoming digitally endangered \cite{rehm2023equality}. 
An output of the survey revision is that the research community generally assumes the use of English as the ``lingua franca'' in NLG tasks. A clear example of this is that there is no mention of the language chosen for the datasets in most surveys, from which we can infer that they are using the English language.  
Indeed, emergent multilingual research takes English as the ``pivot'' language through which models are tested in other more low-resourced languages, as in machine translation, text summarization, etc. However, one of the risks of this methodology is missing some of the semantic properties inherent to each particular language if we take as a basis only one language. This also creates the problem of making it difficult to generalize in NLG architectures if we always copy models from the same language. Thus, extended approaches with each language as the central element of the architecture need to be addressed in future studies. Such multilingual context would also be necessary to study variances between languages and check if NLG models achieve the same performance. Linguistic structures differ between languages, therefore, further research needs to verify if the datasets with which the models are trained are as suitable for other languages as they are for English. 

Another drawback found from the analysis is that even high-resourced languages still lack original datasets in some of the most well-known NLP tasks. Regarding Spanish, it is currently considered the second most spoken language by native speakers in the world, and the third language most used on the Internet after English and Chinese\footnote{Facts extracted from the online report: https://shorturl.at/gCLUZ} (which are the most used languages in the surveys analyzed). 
Nevertheless, most datasets available on NLG specialized websites such as HuggingFace\footnote{\url{https://huggingface.co/datasets}} are (semi)automatic translations of English instead of considering Spanish semantic nuances as the scope of research. Consequently, another research gap in most current NLG surveys is the need for NLG systems oriented to high and low-resourced languages other than English.

\begin{figure}[H]
\centering
\includegraphics[width=0.7\textwidth]{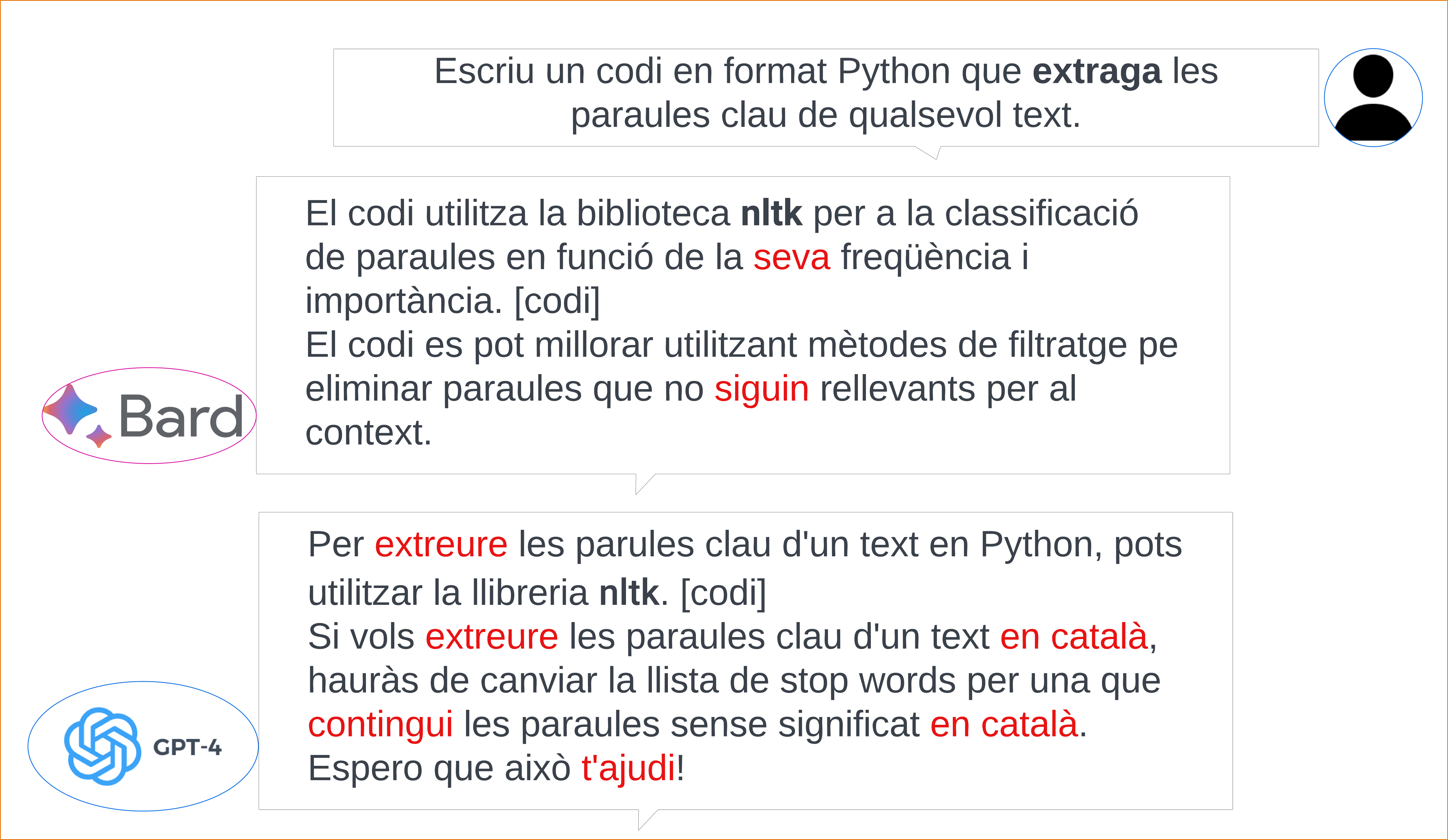}  
\caption{Example of a chat with Bard and GPT-4 in Valencian language.}
\label{fig_multilingual}
\end{figure}

These difficulties when addressing NLG tasks for low-resourced languages are also reflected in the test we performed with both GPT-4 and Bard. For this test, we chose a variant of Catalan, Valencian, as the low-resourced language with which we would try to communicate with both models. This language, which is spoken in the Autonomous Community of Valencia (\textit{Communitat Valenciana}) in Spain, shows very similar linguistic structures to those from the Catalan language, although well-differentiated grammatical exist that make it necessary to create specific language models that raise awareness of the importance of including linguistic variants as this one. As Figure \ref{fig_multilingual} shows, we asked both GPT-4 and Bard a simple command in Valencian (translated here for clarity purposes): ``Write a piece of code in Python format that extracts keywords from any text.''. For this command, we included a Valencian verbal tense (\textit{extraga} --- extracts) that differs in form with respect to the same verbal tense in Catalan (\textit{extregui / extragui} --- extracts), in order to check if the model is capable of automatically recognizing the language without providing further information about the language in which the message is written. 
Both answers were coherent and showed correct linguistic style and grammar. 
However, GPT-4 and Bard tend to get confused with Catalan variants when it comes to the verb ending of the Valencian subjunctive form as in \textit{siguin}, \textit{contingui}, \textit{ajudi} (``will be'', ``will have'' and ``will help''), which should be written as \textit{siguen}, \textit{continga}, and \textit{ajude}. Moreover, a clear example of how Bard interprets our command as in Catalan is the use of the Catalan variant for the possessive pronoun \textit{seva} (``its'') instead of \textit{seua}. In the same vein, GPT-4 directly takes for granted that we are asking for a Catalan response even if we wrote the command in Valencian, as part of its response says: ``If you want to extract keywords from a Catalan text...''. With these examples, we can conclude that further efforts need to be made with LLMs so they can become inclusive tools which can communicate in the wide variety of languages (either high or low-resourced) humans can speak. By doing this, we would avoid the exclusion of any community from using such powerful tools because of their language.

\subsection{Knowledge Integration and Controllable NLG}

Neural Models trained exclusively on a specific type of data, whether multimodal or not, possess constrained knowledge for generating the desired text. Including additional knowledge in neural models could enhance their performance and thereby, obtain a satisfactory output. 
Knowledge can be extracted from two different sources, internal and external \cite{yu2022textgeneration}. The former is obtained from the input text, such as keywords or linguistic features, and the latter is the knowledge that comes from outside sources, such as knowledge bases or external knowledge graphs.

Many studies have focused on two key steps involved in effectively integrating knowledge. The first step is concerned with obtaining helpful knowledge from different sources, and discarding what is irrelevant. The second step focuses on the successful understanding of knowledge and its incorporation into neural models. 
In our survey review, we have identified that despite recent efforts which have contributed to significant progress in this area, there are still several gaps when it comes to effectively integrating knowledge in neural models. 

Regarding controllable NLG, this topic arises from the need to control the final attributes of a text. The generation is guided by a control condition that can be, for example, stylistic (e.g., emotion or order of a text), or it could include some specific content (e.g., keywords or entities), as well as being based on demographic attributes of the speaker \cite{zhang2022survey}). 
There are two promising research lines around this topic: (1) to propose a unified framework to address the controllable generation task. Most of the research in this area has focused on a specific task with specific conditions, so there lacks a global and unified framework. (2) To include additional commonsense knowledge to make models generate texts according to a certain degree of fiction depending on the typology of the output, e.g. the degree of commonsense needed to write a tale varies from the degree of commonsense needed to write a news article. 

\begin{figure}[h]
\centering
\includegraphics[width=0.6\textwidth]{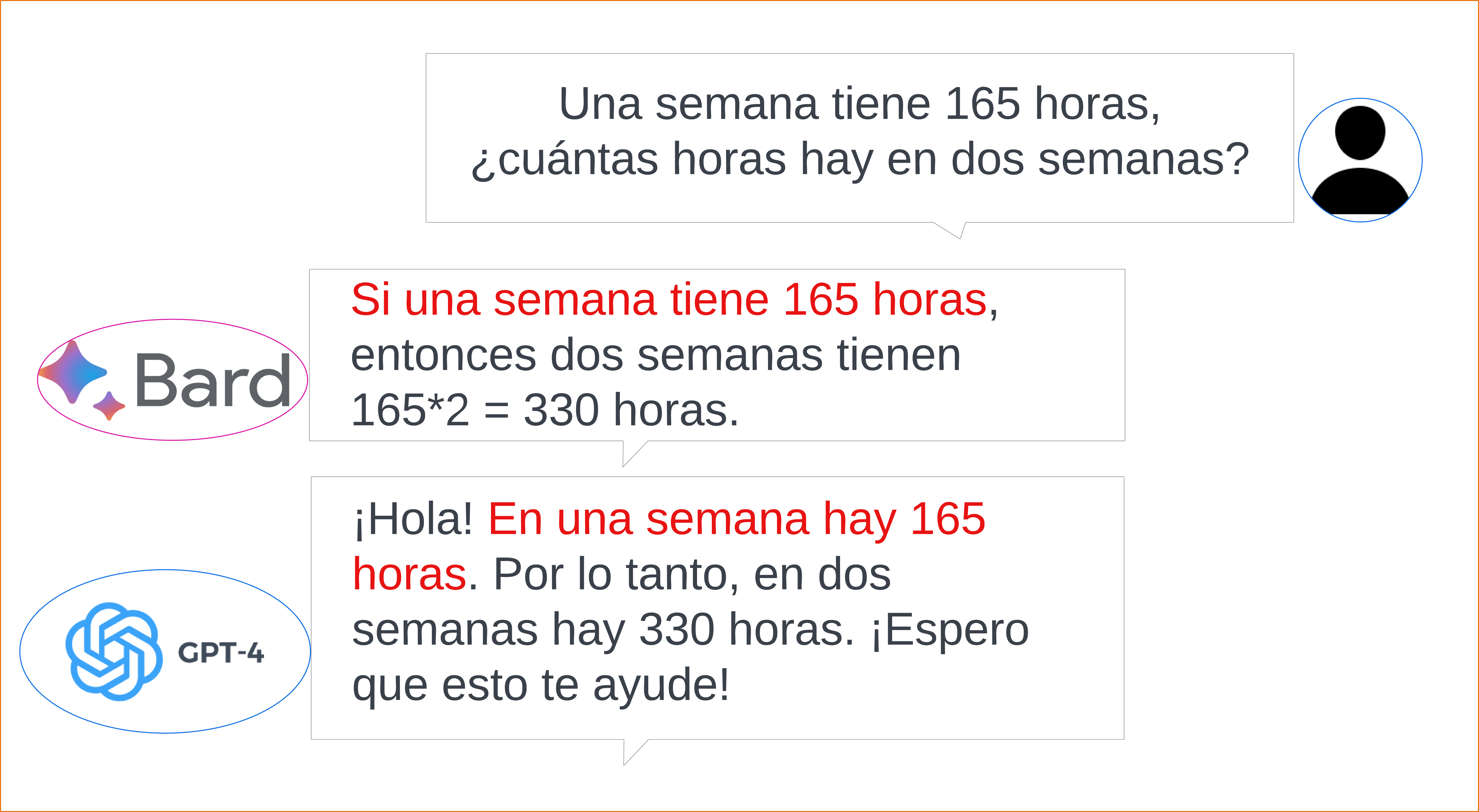}  
\caption{Example of knowledge integration in both LLMs.}
\label{fig_knowledge}
\end{figure}

The limited degree of commonsense that current LLMs show can be easily detected with a simple query in either Bard or GPT-4, to name but a few models. Indeed, they tend to take for granted the information that the user provides in the first input of the chat, and considering that information as true, they develop their answers presuming that the first message only contains real factual data. Figure \ref{fig_knowledge} serves as an example of this interaction with a LLM. In this conversation, we asked both Bard and GPT-4 a simple question taking for granted a factual datum which is actually incorrect. Our first message, ``A week has 165 hours, how many hours are in two weeks?'' implies the incorrect fact that a week consists of 165 hours, being 168 the correct amount. However, both models take for granted this information, and even GPT-4 answers by reaffirming the incorrect statement (``Hi! A week has 165 hours. Therefore, two weeks are 330 hours.''). Consequently, further research needs to be performed so that future LLMs are capable of detecting these factual inconsistencies from the beginning by adding further world knowledge into NLG models. In this manner, LLMs would be able to automatically correct wrong statements so the response they give matches reality and thus provide correct information.

\begin{figure}[!]
\centering
\includegraphics[width=0.6\textwidth]{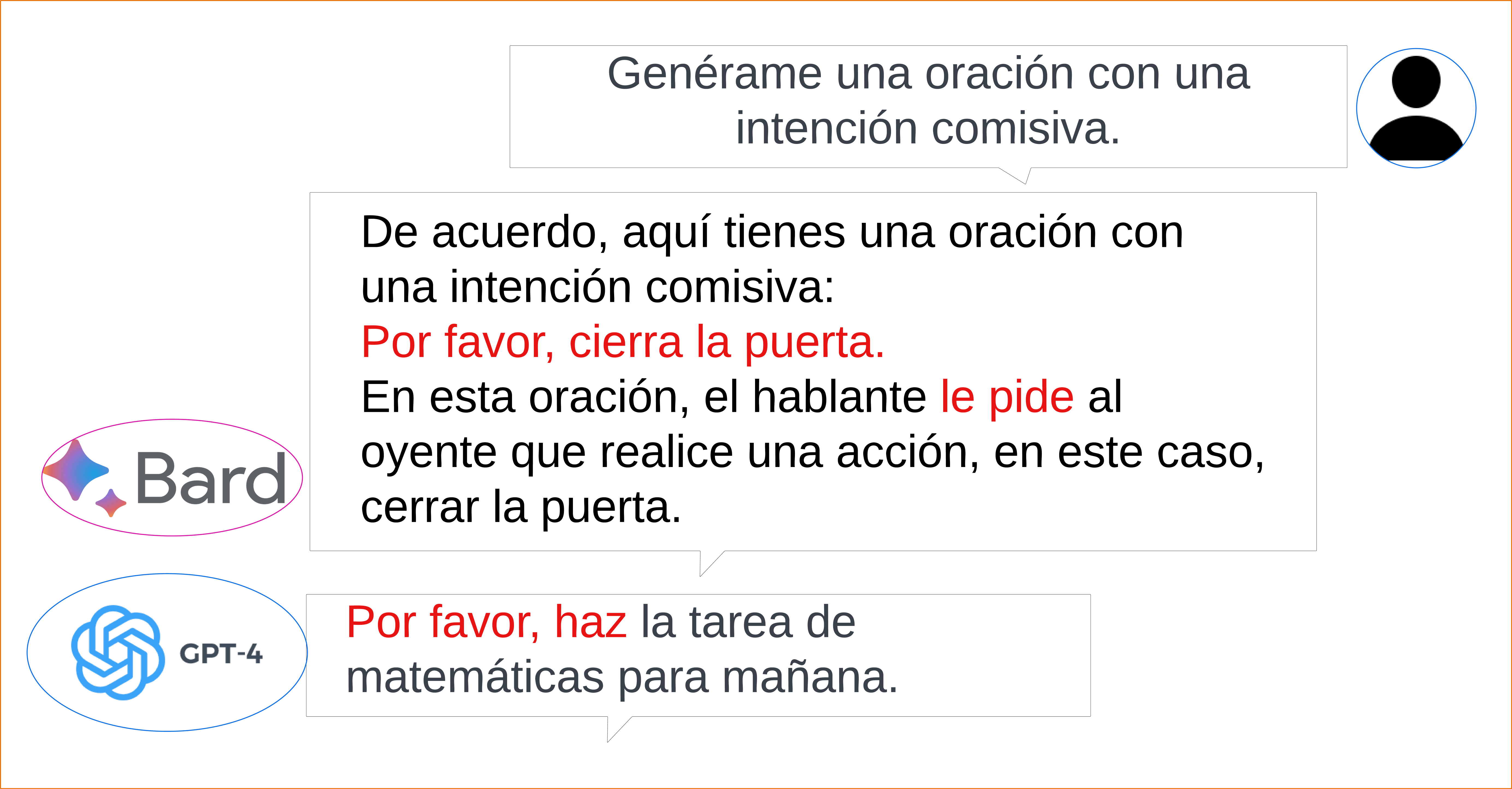}  
\caption{Example of controlled text generation with both LLMs.}
\label{fig_controllable}
\end{figure}

As for controlled NLG, we focused on the task of text generation given a particular communicative intention. In our interaction with GPT-4 and Bard in Figure \ref{fig_controllable}, we requested both models to generate a sentence with a ``commissive'' intention, one of the five types of intentions included in the well-known Speech Act Theory by \cite{austin1962things,searle1969speech}, which is equivalent to making a promise. Surprisingly, both models got confused with the type of intention we wanted the generated message to have, as they generated a sentence with a ``directive'' intention, which encompasses those intentions referred to orders, suggestions and recommendations, among others. The generated sentences, ``Please, close the door.'' (Bard) and ``Please, do your mathematics homework for tomorrow.'' (GPT-4) clearly reflect this intention by using the imperative form of both verbs, which is one of the main indicators of an order and, therefore, a directive intention. Thus, controllable NLG still has a lot of room for improvement, and even more, if we consider pragmatic aspects of language (Spanish in our case). By improving this type of issue, LLMs would be able to, on the one hand, fulfill generation tasks with controlled language specifications and, on the other hand, address the linguistic level of pragmatics, which is still set aside in most tasks within this subdiscipline.

\subsection{Hallucination}
Hallucination is an issue present in state-of-the-art NLG tools. It occurs when a generated text seems to be fluent and natural, but its content is untrustworthy or illogical \cite{ji2023survey}. Hallucinations can be intrinsic when a generated output differs from the source content, and extrinsic when a generated text cannot be corroborated by searching in the source text. Their origin can stem from two primary sources: 
\begin{itemize}
    \item {\it Data}---State-of-the-art models need huge amounts of data to be trained. When building the datasets needed to train these models, some contradictions between the source and target can be introduced and consequently favor the appearance of hallucinations. Another problem is that duplicated data could bias the model to generate repeated data with more frequency.
    
    \item {\it Training and inference}---An inadequate training strategy can also introduce hallucinations. On the one hand, an encoder with a feeble understanding ability could learn wrong correlations of the training data. On the other hand, a decoder could focus on an erroneous part of the encoded input data, leading to hallucinations. 
    Finally, the decoding strategy is also important because a strategy that increases the diversity of the generated output also increases the likelihood of hallucinations \cite{ji2023survey}.

\end{itemize}
The hallucination issue has emerged rapidly as one of the key problems for NLG tools, especially prevalent in the popular LLMs. An unethical use of these tools that exploit the threat of hallucination could potentially be used to generate dis- and mis-information. 
To deal with this, the NLG field has proposed methods and models that could reduce the hallucination problem, although there is still room for improvement. 

\begin{figure}[h]
\centering
\includegraphics[width=0.7\textwidth]{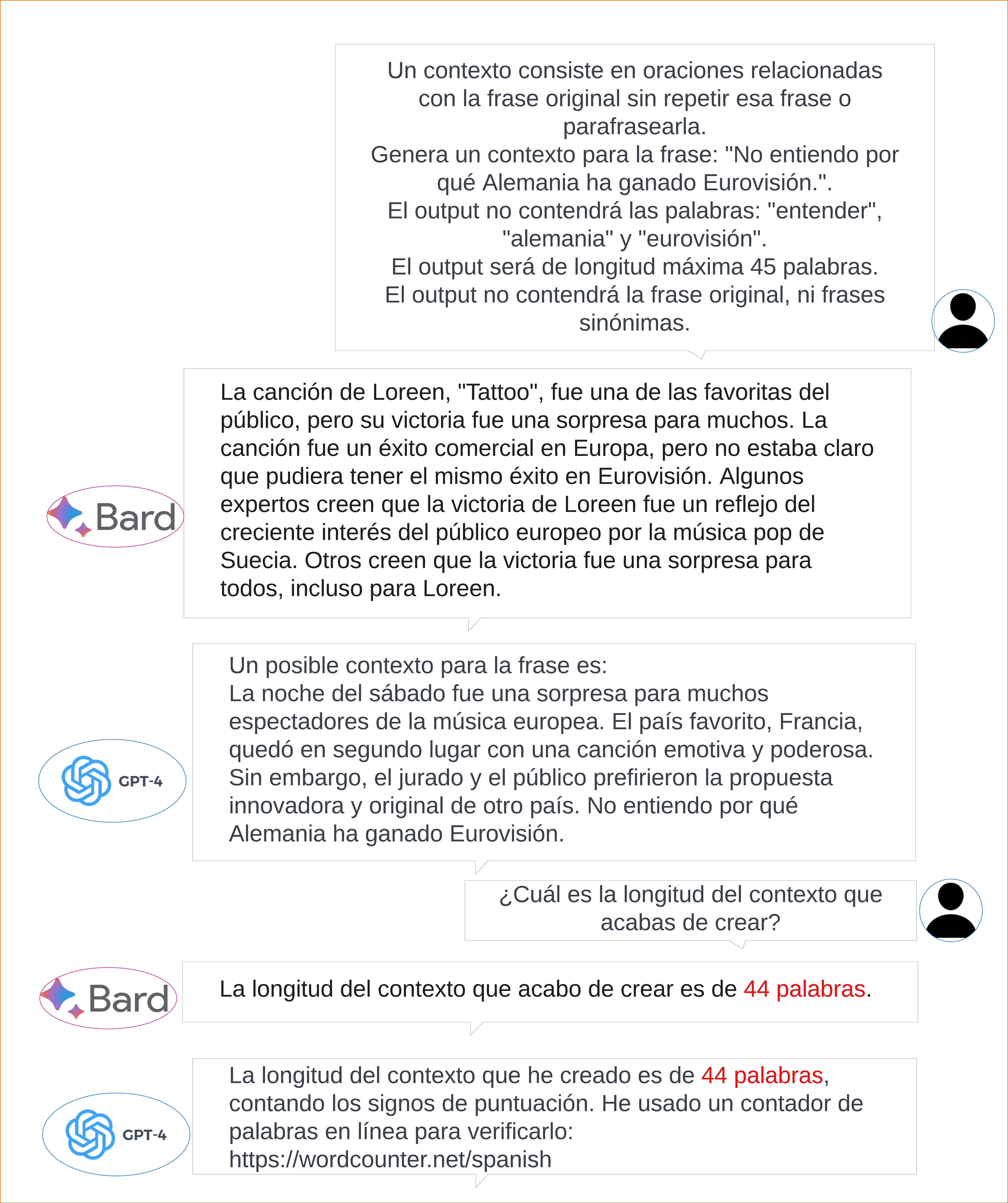}  
\caption{Example of a generated hallucination in both chats.}
\label{fig_hallucination}
\end{figure}

Indeed, many examples of hallucination can be easily detected when communicating with LLMs, and both Bard and GPT-4 are not an exception. For this test, illustrated in Figure \ref{fig_hallucination}, we focused on the task of contextual generation. 
In this task, selected LLMs had to generate a context inspired by an original sentence and three keywords extracted from that sentence which could not be included in the generated context, apart from other restrictions determined in the prompt we used for each model. We wrote a prompt with these instructions for both Bard and GPT-4, and detected that, even though the context was generated appropriately, 
both models answered incorrectly when asking them for the length of the context they generated, as both coincided that the length of the generated text is 44 words, although the correct length is 52. In fact, in GPT-4's response, the model even confirms its answer by including the source from which it extracted the ``right'' length of the generated context, although when manually checking the length through the same online counter, the length is indeed 52. The answer translated into English says ``The length of the context I created is 44 words including punctuation marks. I used an online word counter to check it: https://wordcounter.net/spanish''. With this example, we can arguably say that current LLMs need to strengthen their architectures so they are able to double-check the information they provide to users. This need arises from the tendency of LLMs to generate text that may seem logical on the surface, but when manually checking the facts they provide, they may come up with hallucinations like these, making them untrustworthy information resources.

\section{A Step Beyond: New Considerations Triggered by Generative AI}

The exponential growth that Generative AI methods have shown in the last few years has also brought a window of opportunities for researchers in the NLG discipline. They have focused on addressing new tasks and solving already explored areas from new perspectives with a view to boosting the performance of existing models. Indeed, LLMs have proven to be good architectures for the general NLG tasks mentioned in this paper, but with their great performance, new factors have emerged that need to be taken into account. Considering this, the NLG survey review also enabled us to determine which lines of future research should be addressed to confirm our idea that the NLG discipline still has a lot of work to do in the AI field.

\subsection{Explainability}
Deep neural models, such as LLMs, have improved the effectiveness of NLG. Notwithstanding, these techniques have led indirectly to another social concern, which is explainability \cite{xu2019explainable}. Traditionally, NLG models were seen as \textit{white box} systems where the decisions made by the models were guided by rules or decision trees. Consequently, these systems were inherently explainable. Since the development of deep neural models, improvements in performance have come at the cost of interpretability.  These models, seen as \textit{black box} systems, produce an output with no explanation of why the model has selected that result, or why it has arrived at a specific decision. As a result, it may trigger a lack of trust among users of these systems.

For these reasons, Explainable AI has become an interesting topic for the research community, and specifically the NLG field, to address. 
Ensuring that a system provides transparency as to how it arrives at decisions could help developers and users of systems. Explainability could help developers to detect data bias, identify mistakes made by the models, such as hallucinations, and improve these flaws. End users can also benefit when a system provides a decision as output because end users can understand why the system arrived at a decision and evaluate the trustworthiness of the steps taken. In this way, mistakes in reasoning can also be identified. Finally, explainability can be crucial in different socially impactful fields such as finance, medicine, or marketing.   
To sum up, although some advances have been made in this area, we still need more trustworthy and transparent NLG systems.

\subsection{Narratives that Engage}

LLMs 
are able not only to generate narrative texts, but also to generate them with creativity. Essentially, they are able to create new stories from scratch, with characters, time-ordered events, dialogues, etc.\cite{alabdulkarim2021}. 
However, there are other aspects of narrative texts that LLMs are not able to deal with, mainly because it is not possible to model these narrative components with only contextual relationships between words and sequential generation. Therefore, other complementary approaches to narrative generation are necessary to improve these aspects as follows:

    \begin{itemize}
    \item {\it Coherence}---Narrative structures are based not only on time-related events, but also on causality. Narrative coherence is based on the cause-effect relationship between events, but LLMs are not capable of representing or generating these causal relationships \cite{alabdulkarim2021}.
    \item{\it Plot}---It is assumed that all narration must be interesting and that happens when there is a conflict and (possibly) a final resolution \cite{AlhussainAzmi2021}. However, LLMs do not have an overview of the narrative to create a well-planned plot with interest for the reader.
    \item{\it Suspense}---A special feature of the narrative to capture the reader's attention is to generate suspense. This implies control of what information is shown to the reader, where their attention is focused, the horizon of expectations and (if applicable) the breaking of those expectations, etc. None of these aspects are considered by LLMs. As a result, they generate boring narratives \cite{alabdulkarim2021, AlhussainAzmi2021}.
    \item{\it Characters}---One of the most relevant components of narratives is the characters. The psychological depth and authenticity of the characters arguably trigger reader connection and empathy. 
    Moreover, narrations depend heavily on character relationships. Of course, LLMs struggle to develop interesting, authentic and thereby relatable characters or the relationships between them \cite{alabdulkarim2021}.
\end{itemize}


These four aspects of automatic narrative generation that LLMs are unable to manage are, therefore, open research topics in NLG that need complementary models. For some of these aspects, controlled generation \cite{alabdulkarim2021,kybartas2016survey} is necessary, where a human decides how the narrative should be created.

\subsection{Prompt Engineering and Beyond}

Prompt engineering is the practice of optimizing textual input for generative AI \cite{white2023prompt}. However, the flurry of interest in this field may not have a lasting impact, according to \cite{acar2023prompt}. The reason behind this is that as AI systems become more intuitive in understanding natural language, the need for meticulously crafted prompts is expected to decrease. New AI language models like GPT-4 also show promising results in generating effective prompts when asked, potentially rendering prompt engineering obsolete. Moreover, the effectiveness of prompts is often limited to specific algorithms, making them less universally applicable across different AI models and versions.
As argued in \cite{acar2023prompt}, problem formulation is a more enduring and adaptable skill for leveraging the potential of generative AI. Problem formulation involves identifying, analyzing, and delineating problems. Well-formulated problems are crucial for achieving effective solutions, even when using sophisticated prompts. However, problem formulation is often overlooked and underdeveloped, with a disproportionate emphasis on problem-solving rather than problem formulation. Following \cite{acar2023prompt}, four key components of effective problem formulation are highlighted: diagnosis, decomposition, reframing, and constraint design. While prompt engineering is currently on everyone's radar, its lack of sustainability, versatility, and transferability restrict its long-term relevance. Emphasizing problem formulation over perfecting prompts enables a better understanding of problems and fosters effective collaboration with AI systems. Bearing this in mind, NLG could also consider wider approaches based on problem formulation which provide a platform for incorporating external knowledge and commonsense into generative AI.

\subsection{Efficiency Issues}


As reported by \cite{trabelsi2021neural}, one of the disadvantages of LLMs is their high computation cost, causing constraints for both training and inference. This means a processing limit on text length, as well as limits on access to updated data (e.g. ChatGPT’s training data only goes up to 2021), which could be a serious handicap especially in NLG tasks. For example, LLMs have been successfully applied to Open-Domain Question Answering by generating answers to users’ queries. However, the previous phase of compiling the passages of the relevant documents to extract the answer implies ad-hoc document retrieval, which is limited to the necessary processing of longer documents than LLMs allow (e.g. BERT cannot take input sequences longer than 512 tokens). In this way, the training of the LLM for this task is usually formed by triples such as “[document [CLS], query [SEP], passages [SEP]]” that frequently exceed 512 tokens.

To overcome this issue, several proposals have been developed, in which computational cost and memory complexity plays an important role. The common solution is to split the documents into smaller pieces of text, whether sentences or passages. However, as stated in \cite{kitaev2020reformer}, ranking documents of length “L” using Transformers can require \(\mathcal{O} (L\textsuperscript{2})\) memory and time complexity (the authors reduce this complexity to \(\mathcal{O}(L \cdot \log L))\), which renders these solutions unfeasible, even though the extraction of LLM-based document representation are run offline.

Therefore, decreasing memory complexity is an important research line in this area. For example, to reduce the dimension of the embeddings, vector compression methods have been proposed. Likewise, the combination of traditional bag-of-words (BOW) approaches (e.g. BM25) that filter the set of documents to a reduced set of passages, which are reranked using LLM-based semantic and relevance modes. Some researchers advocate for discarding these BOW approaches because they do not contain lots of important semantic information about documents. Thus, by proposing LLM embeddings to perform efficient retrieval based on the product quantization technique will assign for every document a real-valued codeword from the codebook or a binary code as in semantic hashing.

\subsection{Ethical concerns}
LLMs can be a powerful tool to help humans in their daily life activities when used responsibly. However, given the large scale these models have acquired with their latest developments, several ethical considerations have emerged to preserve users' integrity, personal privacy and at the same time mitigate the wide range of societal biases that LLMs may reflect, which can come from very different sources \cite{hovy2021bias}. 
Indeed, LLMs' potential has made researchers test their performance in increasingly specific tasks across professional disciplines which are not exempt from controversial decisions with serious consequences for humans. Within the legal setting, the paper published by \cite{chen-etal-2019-charge} raised a discussion about the limits of using NLP tools for legal decisions, as this work focused on the automatic prediction of prison terms via a dataset of records published by the Supreme People's Court of China. 
As for clinical NLG, the accuracy of the predictions that NLG architectures may provide cannot leave room for any mistake or doubt, as their generated information can have severe consequences for the patients those results are directed to. At the same time, legal concerns need to be considered within this professional field, as many studies need to feed their models with patients' medical records in order to learn clinical predictions, although by getting such data they may run the risk of interfering with the personal privacy of patients \cite{Thirunavukarasum2023clinical}.  
Another current issue LLMs are coming up with is the existence of gender bias in either the data those models are trained with or, as a consequence, in the information generated by those models. Language is a reflection of society, and when LLMs reflect these societal biases, they perpetuate harmful stereotypes for people belonging to different social groups \cite{vashishtha2023bias}. NLP research has already addressed the societal biases automatically reproduced in linguistic processing systems. Unfortunately, very little work has been done when approaching gender bias from the NLG perspective \cite{garimella2021bias}. Given this lack of research on language generation biases, we believe that it is of high importance to address this issue by also considering the several grammatical structures used in different languages for detecting biases. The reason for this is that languages differ in the structures used to express a particular human genre given their cultural and societal context \cite{vashishtha2023bias}, so different approaches would have to be tested to mitigate this NLP issue. 
In this work, we only mentioned some of the tasks in which ethical issues may come up when automatically processing information, but these same concerns could be applied to many other research fields, as it has already been done in the area of news processing and how they deal with dis- and mis-information \cite{dulhanty2019disinformation}, as well as the ethical consequences of using crowdworkers to so labelling and evaluation tasks within NLP research \cite{shmueli2021crowdsourcing}. In summary, such is the awareness of the ethical considerations that NLP researchers need to include in their work that the European Union already published the document ``Ethics Guidelines for Trustworthy Artificial Intelligence'' in 2019 \cite{EU2019ethics}. This document, which includes sections devoted to both the creation and the evaluation of trustworthy AI, serves as a model of how researchers should develop new technologies to preserve human integrity and mitigate untrustworthy information. Consequently, we believe that future studies focused on the creation of models with an ethical approach will be beneficial for both the NLP research community and society so we can benefit from their potential preserving their trust.

\section{Conclusions}

This survey outlined the state of the art in NLG by analyzing current key research lines as well as other forward-looking promising ones, derived from the gaps identified in the survey review. An analysis of 16 of the most recent surveys published in the field identified the crucial areas that are being addressed in the context of NLG. The analysis also sheds some light on other unsolved and important problems to tackle. Indeed, although Generative AI and LLMs are capable of solving many NLG tasks by following a one-fits-all approach, they still have a lot of room for improvement to generate reliable and top quality texts. We consider that this survey can help researchers in the NLG field to identify potential research topics to address and draw a roadmap that guides NLG along its future path.

The resulting roadmap for future research lines within the NLG field focuses on the present research gaps concerning LLMs in the areas of multimodality, multilinguality, knowledge integration and controllable NLG, as well as hallucination.   
Moreover, identifying such gaps has enabled us to determine future research lines resulting from the evolution of LLMs. These include the following areas: LLMs explainability; creating engaging narratives; looking beyond prompt engineering; the efficiency of such models and, several ethical concerns when using LLMs. We consider that this work can help researchers in the NLG field to identify potential research topics to address and draw a roadmap that guides NLG along its future path.

\section*{Acknowledgments}
This research is part of the R\&D projects ``CORTEX: Conscious Text Generation'' (PID2021-123956OB-I00), funded by MCIN/AEI/10.13039/501100011033/ and by ``ERDF A way of making Europe''; ``CLEAR.TEXT: Enhancing the modernization public sector organizations by deploying Natural Language Processing to make their digital content CLEARER to those with cognitive disabilities'' (TED2021-130707B-I00), funded by MCIN/AEI/10.13039/501100011033 and ``European Union NextGenerationEU/PRTR''; and project ``NL4DISMIS: Natural Language Technologies for dealing with dis- and misinformation with grant reference (CIPROM/2021/21)" funded by the Generalitat Valenciana.  Moreover, it has been also partially funded by the Ministry of Economic Affairs and Digital Transformation and ``European Union NextGenerationEU/PRTR'' through the "ILENIA" project (grant number 2022/TL22/00215337) and "VIVES" subproject (grant number 2022/TL22/00215334). 

\end{document}